\documentclass[conference]{IEEEtran}
\pdfoutput=1
\IEEEoverridecommandlockouts
\usepackage{cite}
\usepackage{amsmath,amssymb,amsfonts}
\usepackage{algorithmic}
\usepackage{textcomp}
\usepackage{xcolor}
\usepackage[pdftex]{graphicx}
\usepackage{epstopdf}
\usepackage{lipsum}
\usepackage{graphicx, type1cm, lettrine}
\def\BibTeX{{\rm B\kern-.05em{\sc i\kern-.025em b}\kern-.08em
    T\kern-.1667em\lower.7ex\hbox{E}\kern-.125emX}}
\begin{document}

\title{A Deep Learning Approach To Dead-Reckoning Navigation For Autonomous Underwater Vehicles With Limited Sensor Payloads\\
}
 
\author{\IEEEauthorblockN{Ivar Bj\o rgo Saksvik}
\IEEEauthorblockA{\textit{Department of Mechanical, Electronic} \\ \textit{and Chemical Engineering} \\
\textit{Oslo Metropolitan University}\\
Oslo, Norway \\
s309664@oslomet.no} 
\and
\IEEEauthorblockN{Alex Alcocer}
\IEEEauthorblockA{\textit{Department of Mechanical, Electronic} \\\textit{and Chemical Engineering} \\
\textit{Oslo Metropolitan University}\\
Oslo, Norway \\
alepen@oslomet.no}
\and
\IEEEauthorblockN{Vahid Hassani}
\IEEEauthorblockA{\textit{Department of Mechanical, Electronic} \\ \textit{and Chemical Engineering} \\
\textit{Oslo Metropolitan University}\\
Oslo, Norway \\
vahidhas@oslomet.no}
}
\maketitle
\begin{abstract}
 This paper presents a deep learning approach to aid dead-reckoning (DR) navigation using a limited sensor suite. A Recurrent Neural Network (RNN) was developed to predict the relative horizontal velocities of an Autonomous Underwater Vehicle (AUV) using data from an IMU, pressure sensor, and control inputs. The RNN network is trained using experimental data, where a doppler velocity logger (DVL) provided ground truth velocities. The predictions of the relative velocities were implemented in a dead-reckoning algorithm to approximate north and east positions. The studies in this paper were twofold I) Experimental data from a Long-Range AUV was investigated. Datasets from a series of surveys in Monterey Bay, California (U.S) were used to train and test the RNN network. II) The second study explore datasets generated by a simulated autonomous underwater glider. Environmental variables e.g ocean currents were implemented in the simulation to reflect real ocean conditions. The proposed neural network approach to DR navigation was compared to the on-board navigation system and ground truth simulated positions.
\end{abstract}
\section{Introduction}
\lettrine[lines=3, findent=3pt, nindent=0pt]{A}{utonomous underwater vehicles (AUVs)} have in the last decades become important tools in ocean research. Untethered from umbilical cables, these vehicles are suitable for a high variety of applications including bathymetric mapping, water sampling and environmental monitoring. A notorious challenge for AUVs is to navigate and \textit{georeference} acquired sensor data during operations as GPS signals can't propagate trough water. Conventional solutions to this issue involve adding acoustic navigational or/and positioning instruments to the AUV payload. Due to the good propagation of sound in water, doppler velocity loggers and acoustic baseline systems are considered the backbone in AUV navigation and underwater positioning \cite{b10}, \cite{b35}. However, these traditional sensors are often expensive and consumes large amounts of power. In AUV fleets, the cost of adding acoustic instruments is compounded with the number of vehicles. In this paper we consider a limited sensor suite consisting of an IMU sensor and a pressure transducer, where acoustic instruments are partially available to collect experimental training data. Collected DVL velocity measurements from only a few missions are used as a reference in supervised neural network training. The aim for the trained network is to complement DR navigation when the DVL sensor is inaccessible, for example in AUV fleets with budget limitations.
\newline
\newline
The absence of acoustic navigational and positioning instruments has traditionally been compensated by model-based observers like Extended Kalman Filters (EKFs). These are derived from AUV dynamics to form an estimation model \cite{b8}, \cite{b9}, \cite{b31}, \cite{b32}. Unfortunately, model-based observers rely on parameters that are difficult to obtain in practice. The dynamics of an AUV is derived based on intricate hydrodynamic models. Experiments must be carried out in a towing-tank facility or using expensive CFD (Computational Fluid Dynamics) software to obtain hydrodynamic damping coefficients \cite{b38}, \cite{b3}. If the external geometry of the AUV changes, i.e. when making small modifications to payload sections, the coefficients need to be updated. 
\newline
\newline
To avoid deriving complex AUV models and conducting time consuming towing-tank or CFD experiments, this paper presents a data-driven approach to dead-reckoning navigation. Using experimental data from AUV missions and simulations, a neural network is trained to learn and generalize relative AUV motions. Data-driven neural network regression abolishes the need for knowledge of a dynamic model, and avoids modelling and estimation errors related to classical state observers \cite{b4}, \cite{b5}. A recurrent neural network (RNN) is developed to relax time-delayed effects in the AUV dynamics which occurs due to vehicle inertia, under actuation and added mass effects \cite{b1}, \cite{b6}. With an input layer composed of standard sensory measurements (pressure sensor, inertial measurement unit) and control actions, the RNN network aims to predict relative surge $u_{r}$ and sway $v_{r}$ velocities. These are further implemented in a dead-reckoning algorithm to approximate North and East positions during operations.
\subsection{Related Work}
Several articles have addressed artificial neural network state estimation for marine crafts. In Zhang \textit{et al.} \cite{b4} a Short-Term Long-Term-Memory (LSTM) recurrent neural network is proposed to estimate the relative position of an AUV. The LSTM network used data from a pressure sensor, an inertial measurement unit (IMU), and an acoustic doppler velocity logger (DVL) to predict the horizontal north and east positions. Training and validation data were collected from a series of surface trajectories while logging GPS locations, which were projected as ground truth measurements. A similar study with the same AUV is presented in Mu \textit{et al.} \cite{b5}, where a bi-directional LSTM network was used. A neural network approach to dead-reckoning navigation of dynamically positioned ships is presented in Skulestad \textit{et al.} \cite{b6}. Control actions and commands from vessel thrusters combined with heading measurements was used as input data in a RNN network to aid navigation during GNSS outages. Experiments were conducted in a vessel simulator with time-varying environmental disturbances such as wind forces, sea waves and ocean currents. In Chen \textit{et al.} \cite{b18} a neural network is presented to assist navigation during DVL malfunction. A nonlinear autoregressive network with exogenous SINS (Strapdown Inertial Navigation System) inputs was used. The network was tested and validated on a ship with a DVL mounted on the vessel hull to provide training and validation data.
\newline
\newline
The remaining parts of this paper are detailing the following segments - Section II and III addresses the concept of dead-reckoning navigation and the neural network velocity observer respectively. Section IV presents the AUV platforms and datasets used to train and test the neural networks. The results are detailed in section V and the conclusion and recommendations for further work are presented in VI.
\section{Dead-Reckoning Navigation}
In the absence of GNSS (Global Navigation Satellite Systems) systems, AUVs enters a dead-reckoning mode while under water. The DR algorithm predicts the position of the AUV based on estimates at the previous time-step. With a reference of the heading and attitude combined with relative velocity measurements, the position is determined by numerical integration. To compute the relative position of the AUV, the measured/estimated relative velocities must be rotated with respect to the inertial reference frame of the vehicle. Following \cite{b1} the inertial frame of underwater vehicles is defined by North-East-Down (NED) local tangent plane coordinates. The NED velocities $\boldsymbol{\dot{\chi}} = [\dot{N}, \dot{E}, \dot{D}]^T$ of an AUV are derived by an rotation matrix from the body frame $\{b\}$ to the inertial frame $\{n\}$ \cite{b1}. An AUV influenced by ocean currents $\boldsymbol{\upsilon}_{c}$ will have a relative velocity $\boldsymbol{\upsilon}_{r}$. Assuming that the ocean currents are irrotational they are derived following \cite{b1} as $\boldsymbol{\upsilon}_{r} = [u_{r},v_{r},w_{r}]^T = [u - u_{c}, v - v_{c}, w - w_{c}]^T$. Accordingly, the relationship between the relative body-fixed and inertial velocities are given as 
\begin{equation}
\boldsymbol{\dot{\chi}} = \boldsymbol{R}_{b}^{n}(\boldsymbol{\Theta})\cdot \boldsymbol{\upsilon}_{r}
\label{ned_velocities}
\end{equation}
Where $\boldsymbol{\Theta} = [\phi, \theta, \psi]^T$ is the attitude and heading of the AUV provided by an inertial measurement unit (IMU). Equation \ref{ned_velocities} can be written in expanded form as 
\begin{equation}
\begin{bmatrix}
\dot{N} \\ \\
\dot{E} \\ \\
\dot{D}
\end{bmatrix} = \begin{bmatrix}
u_{r}\cdot c(\psi)c(\theta) + v_{r}\cdot [c(\psi)s(\theta)s(\phi) - s(\psi)c(\phi)] \\ + 
w_{r} \cdot[s(\psi)s(\phi) + c(\psi)c(\phi)s(\theta))] \\ \\
u_{r} \cdot s(\psi) c(\theta) + v_{r} \cdot [c(\psi) c(\phi) + s(\phi) s(\theta) s(\psi)] \\ \\
-u_{r} \cdot s(\theta) + v_{r} \cdot c(\theta) s(\phi) + w_{r} \cdot c(\theta) c(\phi) \\ \\
\end{bmatrix}
\label{DR_expanded}
\end{equation}
where $c() = cos()$ and $ s() = sin()$.
\newline
For AUVs that typically operate with a zero angle of attack the attitude can be neglected in eq. \ref{DR_expanded}. However, for other vehicles like underwater gliders which can perform spiraling motions with non-zero attitude $[\phi,\theta]^T \neq 0$, it persist. After rotating the relative velocities according to the inertial frame of the vehicle, numerical integration is performed to obtain the position. Given the previous predicted position $\boldsymbol{\chi}_{t} = [N_{t}, E_{t}, D_{t}]^T$ the DR algorithm is derived following \cite{b6}
\begin{equation}
\boldsymbol{\chi}(t+1) = \boldsymbol{\chi}(t) + \boldsymbol{R}^{n}_{b}(\boldsymbol{\Theta})\cdot \boldsymbol{\upsilon_{r}}(t+1) \cdot \Delta t
\label{DR_algo}
\end{equation}
Where $\Delta t$ is the time step between the predictions. 
\begin{figure}[h]
    \centering
    \includegraphics[width = 6.4cm]{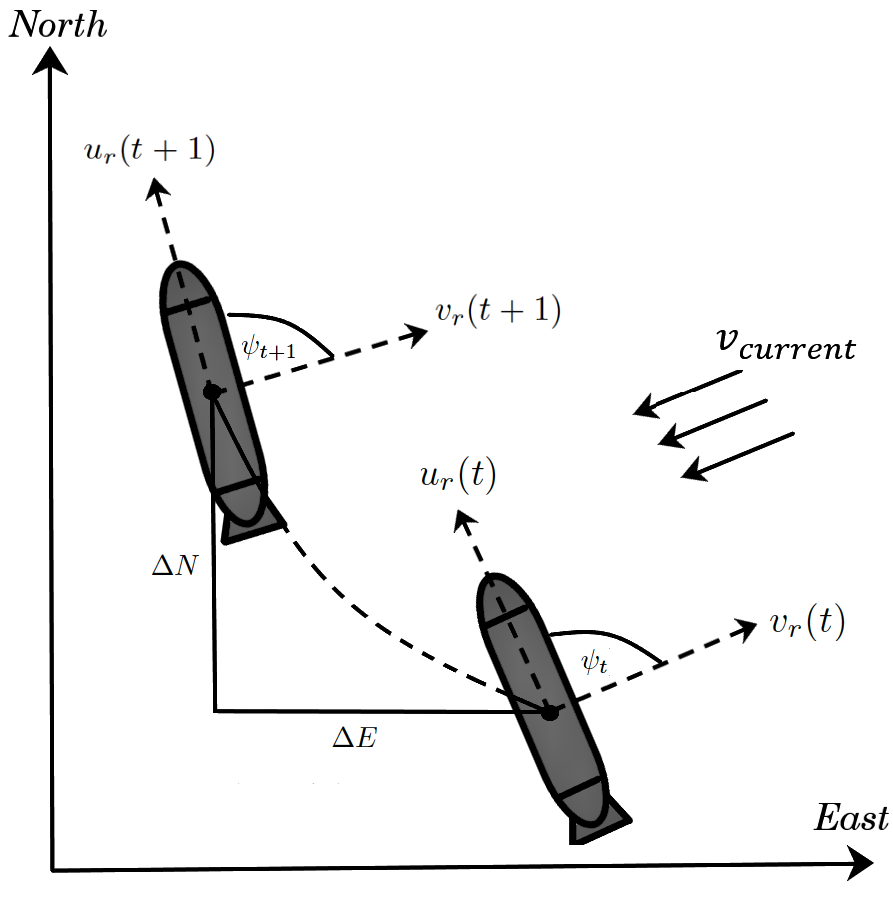}
    \caption{DR navigation Illustration}
    \label{fig:RNN}
\end{figure}
\section{Neural Network Aided Dead-Reckoning Navigation}
\subsection{Sensor Inputs}
On-board sensors like IMUs and pressure transducers contains valuable in-situ information about an AUV. These are used as input variables in the RNN network to predict relative horizontal velocities $[u_{r},v_{r}]^T$.
The development of MEMS (Micro-Electrical-Mechanical-Systems) based IMUs have led to more affordable inertial measurements. They consist of three-axis gyroscopes and accelerometers, typically confined in a silicon chip. The gyroscope, which give measurements of the angular rates $\boldsymbol{\omega^{b}_{IMU}} = [p_{b}, q_{b},r_{b}]^T$ and the accelerometer which provide measurements of the rate-change of velocities $\boldsymbol{\dot{\upsilon}^{b}_{IMU}} = [\dot{u}_{b}, \dot{v}_{b},\dot{w}_{b}]^T$ can be modelled as
\begin{equation}
\begin{split}
\boldsymbol{\omega}^{b}_{IMU} &=  \boldsymbol{\omega}^{b}_{gyro} +  \boldsymbol{b}^{b}_{gyro} \\
\boldsymbol{\dot{\upsilon}}^{b}_{IMU} &=  \boldsymbol{\dot{\upsilon}}^{b}_{acc} +  \boldsymbol{b}^{b}_{acc}
\end{split}
\end{equation}
Where $\boldsymbol{b}^{b}_{gyro}$ and $\boldsymbol{b}^{b}_{acc}$ are sensor biases \cite{b43}. Combined with a three-axis compass, a Kalman Filter derived from a kinematic model impart the euler angles $\boldsymbol{\Theta} = [\phi, \theta, \psi]^T$. A key component in AUVs is the pressure transducer. The relationship between pressure and depth are assumed to be constant, thus the vertical position of the AUV can be approximated by the pressure measurements. Given a measured hydrostatic pressure $\Delta p$, water density $\rho$ and gravitation $g$, the vertical position $z$ and relative heave velocity $w_{r}$ is derived as 
\begin{equation}
z = \rho g \Delta p \implies w_{r} = \boldsymbol{R}^{-1}_{bn}(\boldsymbol{\Theta}) \cdot \dot{z}
\end{equation}
Where $z$ is assumed to be inertial $\{n\}$ and the relative heave velocity $w_{r}$ is represented in the body-fixed frame $\{b\}$ 
\subsection{AUV Control Actuators}
To enforce the neural network state observer, control actions from the AUV actuators are used together with the sensor data. In this paper two AUVs with different actuator configurations are investigated. Conventional underactuated AUVs are normally equipped with an aft thurster and external control surfaces. The thruster is the propulsion system which generates a hydrodynamic force $\tau$ to induce surge transnational motions, while control surfaces consist of external airfoils that alter attitude and heading depending on their deflection angles. The control surfaces typically consist of a rudder and dive planes denoted $\delta_{R}$ and $\delta_{D}$ respectively. Control actions are determined from feedback controllers which in these studies are decoupled into vertical and horizontal manoeuvres. For thruster based AUVs, speed controllers are used to maintain a desired velocity and reject ocean current disturbances.
In addition to the conventional AUV actuators, a variable buoyancy system and internal moving masses are introduced by the AUVs investigated in this paper. The simulated autonomous underwater glider uses buoyancy displacement to alter vertical motions, while a set of fixed wings generates hydrodynamic lift forces to induce forward motions. As gliders operates at low-speeds, control surfaces are ineffective due to low dynamic pressure. Control moments from internal moving masses are used to change the attitude and heading of the vehicle.
\subsection{RNN Architecture}
The RNN architecture is formed with feedback loops in the hidden layers of the network, providing internal \textit{memory} to capture AUV dynamics with time-delays. To improve the estimation of the relative horizontal surge and sway velocities $[u_{r},v_{r}]^T$, two independent neural networks are used for each velocity vector. This is convenient as surge and sway dynamics are often non-interacting or slightly interacting \cite{b1}, \cite{b4}. Using input variables that holds low dependence to the predicted output variables reduces the generalization of the network and may lead to the notorious issue of \textit{overfitting} \cite{b30}, \cite{b24}. 
\begin{figure}[h]
    \centering
    \includegraphics[width = 9.3cm]{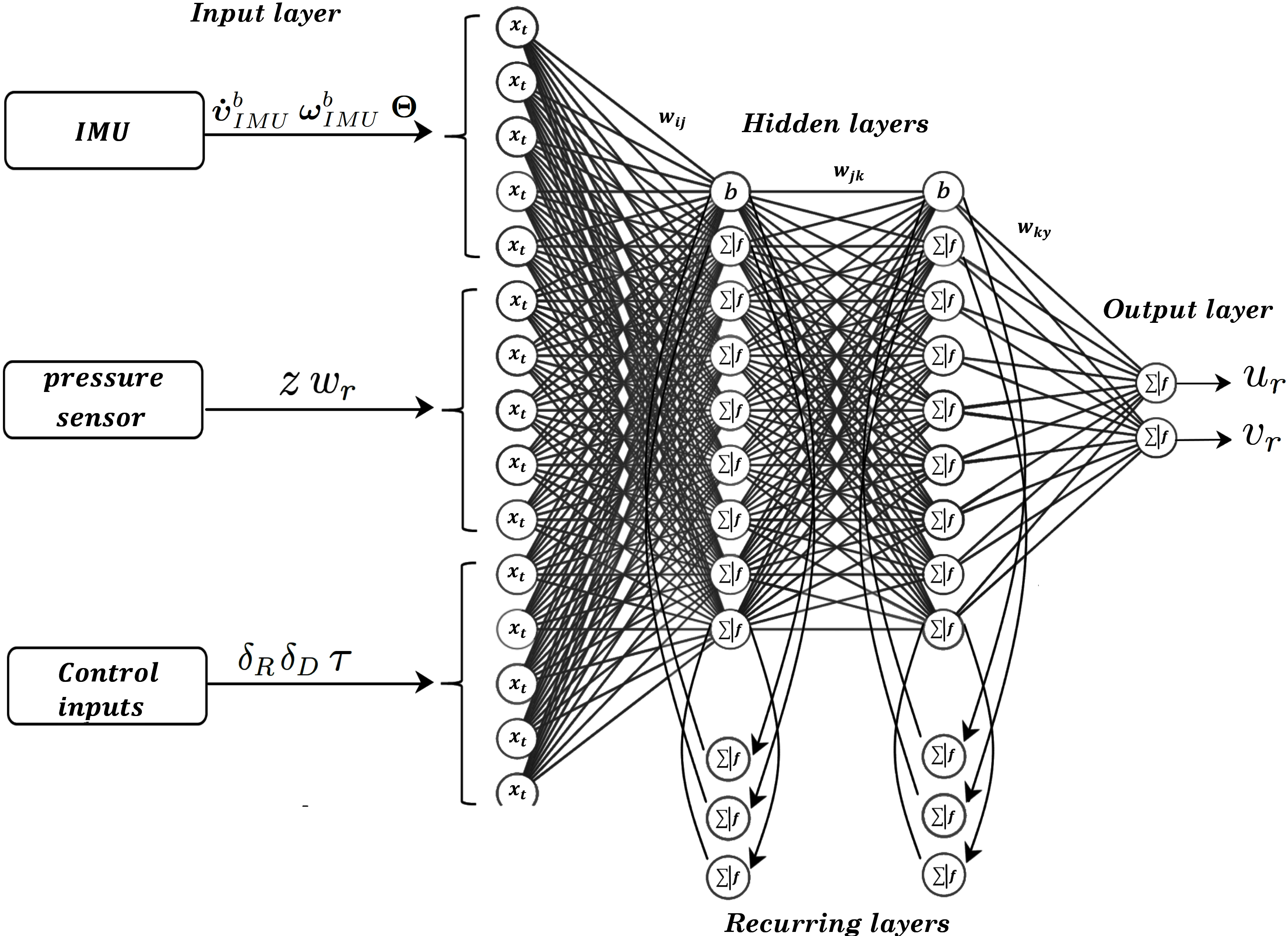}
    \caption{Neural network illustration}
    \label{fig:RNN}
\end{figure}
\newline
RNN networks are generalized feedforward neural networks, but differs as recurring context layers are included in the hidden layers as illustrated in figure \ref{fig:RNN}. The hidden nodes $\boldsymbol{h}(t)$ summarize the inputs $\boldsymbol{x}$ and weights $w_{xh}$ from the previous layer combined with the recurring layer $\boldsymbol{h}(t-i)$. The hidden neurons $\boldsymbol{h(t)}$ and output neurons $\boldsymbol{y(t)}$ is derived mathematically as
\begin{equation}
\begin{split}
\boldsymbol{h}(t) &= \sigma_{h}(\sum w_{xh} \cdot \boldsymbol{x} + \sum w_{hh}\cdot\boldsymbol{h}(t-i) + \boldsymbol{b}_{n}) \\ \\
\boldsymbol{y}(t) &= lin(\sum w_{hy}\cdot \boldsymbol{h}(t) + \boldsymbol{b}_{y})
\end{split}    
\end{equation}
Trough sequential learning based on AUV datasets, the RNN network learns to predict the relative horizontal velocities. Activation functions $\sigma_{h}$ in the hidden layers are the key to learning the nonlinearity between the selected inputs and predicted outputs. The network is trained based on the renowned concept of \textit{backpropogation} introduced in \cite{b34} to tune the weights and biases. The goal of neural network training is to optimize the network parameters so that the error function $E$ is minimized 
\begin{equation}
E = \frac{1}{2} \sum_{i}(\hat{\upsilon}_{r}(i) - \upsilon_{r}(i))^{2} 
\end{equation}  
Where $\hat{\upsilon}_{r}(i) - \upsilon_{r}(i)$ is the difference between the actual and predicted relative velocities.
\subsection{Navigational training data}
In this work, experimental DVL data and simulated velocities are used as a reference for the supervised neural network training. Alternative approaches may involve using acoustic positioning systems which are not prone to cumulative integration errors \cite{b10}, thus providing more accurate ground truth measurements. However, a disadvantage with these approaches is that the DVL/acoustic modem must be replaced with "dummy" sensor to avoid changing the hydrodynamic properties and net weight of the AUV. Ideally, we want to have a reference of the AUV velocities/positions without changing the geometry and weight. A potential solution is to use visual based (machine vision) pose estimation relative to an assisting AUV/ROV.  Machine vision has proven to be successful in autonomous docking operations for AUVs \cite{b39},\cite{b40} and may possibly be extended to tracking applications.      
\section{AUV Platforms $\&$ Datasets}
To train and validate the neural network approach to DR navigation, AUV datasets are needed. In this paper two AUV platforms are investigated. The first dataset originates from sea-trials of the Tethys Long-Range AUV (LRAUV), while the second dataset is from a MATLAB simulation of an underwater glider.
\subsection{Tethys AUV}
\begin{figure}[h]
    \centering
    \includegraphics[width=8.6cm,trim={1.1cm 1.5cm 1.5cm 1cm},clip]{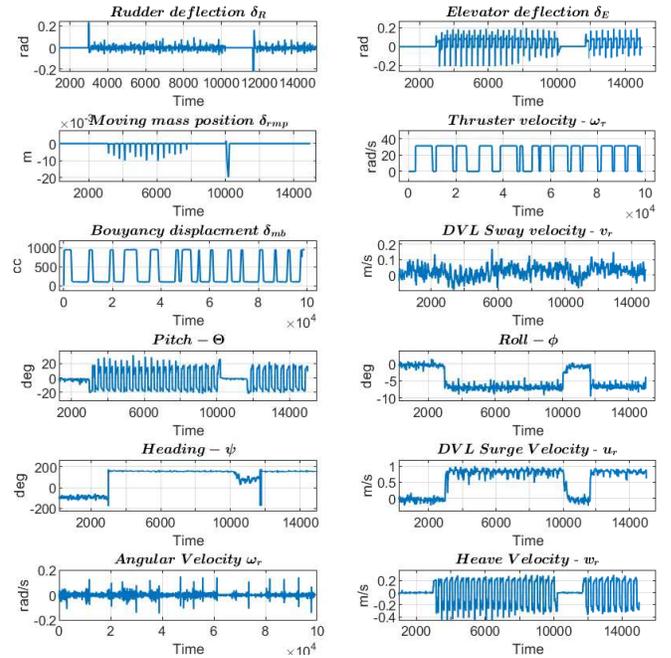}
    \caption{Training data Tethys AUV}
    \label{fig:Tethys_dataset}
\end{figure}

The Tethys Long-Range AUV \cite{b2}, \cite{b20} was developed by the Monterey Bay Research Institute (MBARI) as a research AUV with long-range capabilities. It's characterized as a hybrid AUV as it shares similar control actuators to underwater gliders. This allows it to operate both in undulating glider-like trajectories and at fixed depths. Datasets from the constellation of underwater vehicles at MBARI is available through their public data repository \cite{b24}.
\begin{figure}[h]
    \centering
    \includegraphics[width=5.5cm]{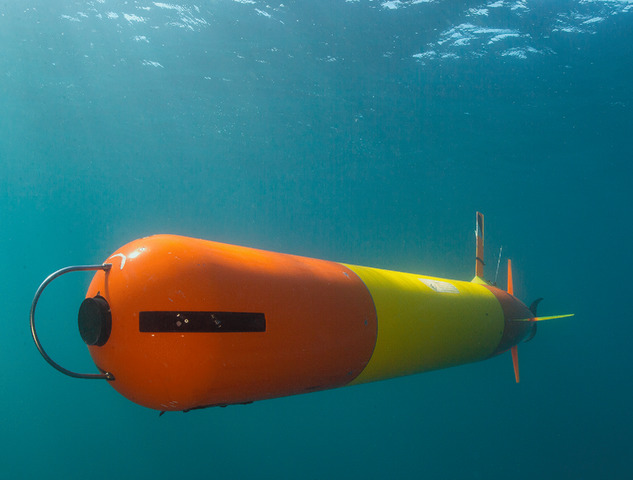}
    \caption{Tethys AUV, courtesy of MBARI}
    \label{fig:Tethys_auv}
\end{figure}
\newline
A series of missions in Monterey bay, California (U.S) were used to train and validate the neural networks. In-situ measurements from a Microstrain 3DM-GX5-24 IMU, Neil Brown pressure sensor and the control actuators were used as inputs to the neural network. Ground truth relative surge and sway velocities $[u_{r}, v_{r}]^T$ were provided by a LinkQuest 600 KHz micro DVL. With periodic GPS fixes the overall navigation accuracy for DVL-aided inertial DR is 3-4 $\%$ \cite{b2}. The navigational errors arise initially from sensor noise and random walk errors from the inertial measurement sensor.
\newline
The parameters for the IMU and DVL sensors hosted on the Tethys is presented in table \ref{tab:AHRS} and \ref{tab:DVL} respectively. The maximum operating altitude refer to \textit{bottom lock} navigation, where the AUV measures it's velocities relative to the seafloor. When out of range the DVL can measure the velocities relative to the water (\textit{water lock}). However, water is often considered as a moving reference frame due to ocean currents, which introduces estimation errors depending on the magnitude of the ocean current vector \cite{b10}. 
 \begin{table}[h]
 \caption{Microstrain 3DM-GX5-25 IMU Parameters}
    \centering
    \begin{tabular}{c|c}
    Error & value   \\
    \hline
    Accelorometer Bias Instability & $\pm$ 0.04 mg \\
    Gyroscope Bias Instability & $\pm$ $8^{\circ}/h$ \\
    Attitude Accuracy & EFK - $\pm$ $0.25^{\circ}$ RMS \\
     Heading Accuracy & EFK - $\pm$ $0.8^{\circ}$ RMS \\
    \end{tabular}
    \label{tab:AHRS}
\end{table}
\begin{table}[h]
 \caption{LinkQuest 600 KHz Micro DVL Parameters}
    \centering
    \begin{tabular}{c|c}
    Parameter & Value \\
 \hline
    Max Altitude & 110 meters  \\
    Min Altitude & 0.3 meters \\
    Accuracy & 1 $\%$ $\pm$ 1 mm/s \\
    Ping rate & 5 Hz \\
    \end{tabular}
    \label{tab:DVL}
\end{table}
\newline
A minor part of the training dataset is presented in figure \ref{fig:Tethys_dataset}. Sensor noise has been filtered out with low-pass filters and Gaussian smoothing. We can observe that the AUV performs saw-tooth trajectories by using the dive plane control surface and internal moving mass actuator. The heading is mostly constant which indicates that the AUV is on a course keeping path governed by a heading controller. 
\newpage
\subsection{Underwater Glider}
The second dataset is gathered from a MATLAB simulation of the Seawing underwater glider \cite{b3}, \cite{b12}. An underwater glider is characterized as a slender body with fixed wings. A variable buoyancy system (VBS) is used to manipulate the volume that the vehicle is displacing to alter vertical motions in the water column. Underwater gliders exploit hydrodynamic properties using fixed wings to generate a forward motion. To control the attitude and heading of the vehicle, an internal moving mass system is used. This typically consists of a battery-pack that can be translated and rotated inside the vehicle housing. A mathematical model of the glider is derived using the 6DOF vectorial marine craft dynamics presented in Fossen \cite{b1}. State variables can be divided into two vectors following SNAME notation \cite{b26} - The position in the inertial frame $\{n\}$ defined as $\boldsymbol{\eta} = [x,y,z,\phi,\theta,\psi]^T$ and the relative velocity in the body frame $\{b\}$ $\boldsymbol{\nu}_{r} = [u_{r},v_{r},w_{r},p,q,r]^T$. Accordingly, a 6DOF kinematic and maneuvering model of an AUV is derived by
\begin{equation}
\begin{split}
&\boldsymbol{\dot{\eta}} = \boldsymbol{J_{\theta}}(\boldsymbol{\eta})\boldsymbol{\nu_{r}} \\ \\
&\boldsymbol{M}\boldsymbol{\dot{\nu_{r}}} + \boldsymbol{C}(\boldsymbol{\nu_{r}})\boldsymbol{\nu}_{r} + \boldsymbol{D}(\boldsymbol{\nu_{r}})\boldsymbol{\nu_{r}} + \boldsymbol{g}(\boldsymbol{\eta}) = \boldsymbol{\tau}
\end{split}    
\end{equation}
Where $\boldsymbol{M} = \boldsymbol{M}_{rb} + \boldsymbol{M}_{A}$ and $\boldsymbol{C}(\boldsymbol{\nu})$ = $\boldsymbol{C}_{rb}(\boldsymbol{\nu})$ + $\boldsymbol{C}_{A}(\boldsymbol{\nu})$ are the transnational and rotational rigid-body dynamics with correlating added mass effects. Hydrodynamic forces and moments are described in the damping matrix $\boldsymbol{D}(\boldsymbol{\nu})$ and the restoring forces are defined by $\boldsymbol{g}(\boldsymbol{\eta})$. $\boldsymbol{\tau}$ is the vector describing the control forces and moments which acts on the vehicle.
\newline
\newline
In presence of ocean currents, the $\boldsymbol{relative}$ $\boldsymbol{velocity}$ is defined by differentiating the body-fixed velocities to the ocean current vector $\boldsymbol{\nu_{r}} = \boldsymbol{\nu} - \boldsymbol{\nu}_{c}$. In the Matlab simulation we consider a two-dimensional irrotational ocean current model. Given an absolute velocity $V_{c} = \sqrt{u_{c}^{2} + v_{c}^{2}}$ we can define the ocean currents in the body frame as
\begin{equation}
\boldsymbol{\nu}^{b}_{c} = 
\begin{bmatrix}
V_{c}\cdot cos( \beta_{c} - \psi) \\
V_{c}\cdot sin( \beta_{c} - \psi) \\
0
\end{bmatrix}
\end{equation}
Many chose to simplify the ocean current model to be constant in the body-fixed frame $\{b\}$, thus $\boldsymbol{\dot{\upsilon^{b}_{c}}}$ = 0 \cite{b27}. However, this only yields during course keeping. A more realistic approach is to assume that the ocean currents are time-varying with respect to rotational motions of the glider. Consider a skew-symmetric matrix $\boldsymbol{S}$ that satisfies $\boldsymbol{S(x)\cdot y} = \boldsymbol{x} \times \boldsymbol{y}$ and the angular velocities in the body frame $\boldsymbol{\omega_{b}}$, the ocean currents can be derived as
\begin{equation}
\dot{\boldsymbol{\nu}^{b}_{c}}= -\boldsymbol{S}(\boldsymbol{\omega_{b}})\cdot \boldsymbol{\upsilon}^{b}_{c}    
\end{equation}
In order to model ocean currents in the 6DOF manoeuvring model, the rotational dynamics detailing Coriolis and centripetal forces must be derived using \textit{velocity-independent parametrizations} \cite{b1}. Given a center of gravity vector relative to the center of origin $\boldsymbol{r}_{cg} = [x_{cg}, y_{cg}, z_{cg}]^T$ the Coriolis and centripetal matrix can be defined as
\begin{equation}
\boldsymbol{C}_{rb}(\boldsymbol{\nu}) = \begin{bmatrix}
m\cdot \boldsymbol{S}(\boldsymbol{\omega_{b}}) & -m\cdot\boldsymbol{S}(\boldsymbol{\omega_{b}}) \cdot \boldsymbol{S}(\boldsymbol{r}_{cg}) \\
m\cdot \boldsymbol{S}(\boldsymbol{r_{cg}})\cdot \boldsymbol{S}(\boldsymbol{\omega_{b}}) & -\boldsymbol{S}(\boldsymbol{I}_{b} \cdot \boldsymbol{\omega_{b}})
\end{bmatrix} 
\label{coriolis}
\end{equation}
Where $\boldsymbol{I}_{b}$ and $m$ is the vehicle inertia and total mass respectively. As demonstrated in eq.\ref{coriolis} the rotational dynamics is derived only using angular velocities $\boldsymbol{\omega_{b}} = [p,q,r]^T$, which satisfies the following property \cite{b1}
\begin{equation}
\boldsymbol{M}_{rb}\boldsymbol{\nu} + \boldsymbol{C}_{rb}(\boldsymbol{\nu}) \boldsymbol{\nu} =  
\boldsymbol{M}_{rb}\boldsymbol{\nu}_{r} + \boldsymbol{C}_{rb}(\boldsymbol{\nu}_{r})\boldsymbol{\nu}_{r} 
\end{equation}
Glider dynamics was simulated in Simulink for a 10-hour interval. Time-varying control inputs and ocean currents were present during the simulation to provide variance in the training dataset. The simulated trajectories of the glider were a combination of undulating wings-levelled motions and turning/spiral manoeuvres. To control the attitude and heading of the glider, two decoupled PID controllers was implemented. The measured state variables were logged and saved to workspace during the simulation. Each training variable holds 81 000 samples, while the validation dataset resulted in 20 000 samples per variable. The test dataset is a combination of wings-levelled manoeuvres and spiral trajectories which differed from the trajectories used in the training data. Hence, we can validate how well the network is generalized to untrained glider motions.
\begin{figure}[h]
    \centering
    \includegraphics[width=8cm,trim={0.68cm 1.6cm 1cm 1.2cm},clip]{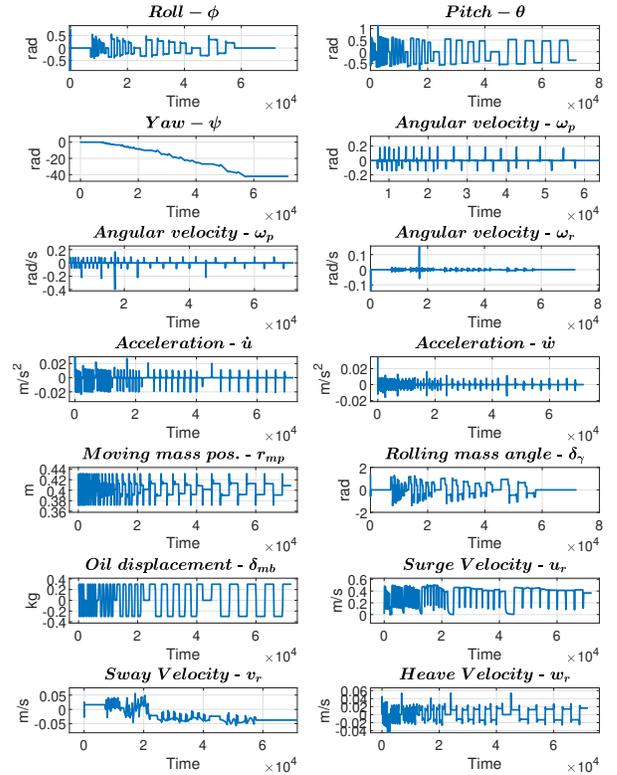}
    \caption{Simulated glider training data}
    \label{fig:seawing_MI}
\end{figure}
\newpage
\section{Experimental and Simulation Results}
This paper presents a two-folded study which consists of simulated and experimental datasets. Collected AUV data was allocated into Matlab and further used to develop, train, and test the neural networks. The predicted outputs was fed into the dead-reckoning algorithm derived in eq. \ref{DR_algo}. 
\newline
The Deep Learning Toolbox was used to design network architectures and perform backpropogation training. A Scaled Conjugate Gradient (SCG) algorithm was chosen as the training function to deal with the large AUV datasets effectively. The SCG algorithm \cite{b34} abolish the need for line-searches as presented in its predecessor \cite{b35} which reduces the computational load. To improve the generalization of the neural networks, \textit{early-stopping} was introduced. Early-stopping divides the AUV dataset into training and validation batches. The training dataset is fed into the SCG algorithm to tune the weights and biases, while the validation data is used to monitor and detect if occurrences of overfitting is evident \cite{b42}. If the network starts to overfit the dataset, the training is aborted, hence the name early-stopping. 
\subsection{Case Study 1 - Tethys AUV}
\begin{figure}[h]
    \centering
    \includegraphics[width=7cm]{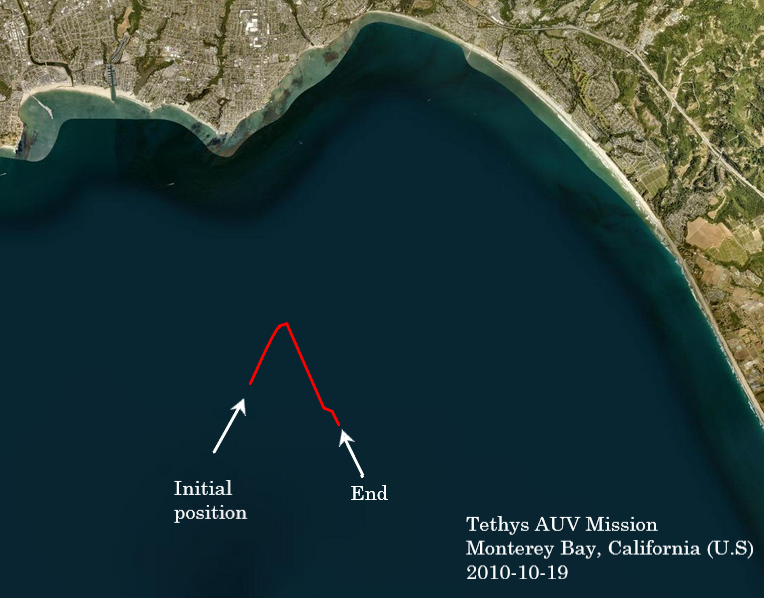}
    \caption{Test trajectory in Monterey Bay, California}
    \label{fig:Tethys_survey_monterey}
\end{figure}

\begin{figure}[h]
    \centering
    \includegraphics[width=7.8cm,trim={0.5cm 0.2cm 0.6cm 0.3cm},clip]{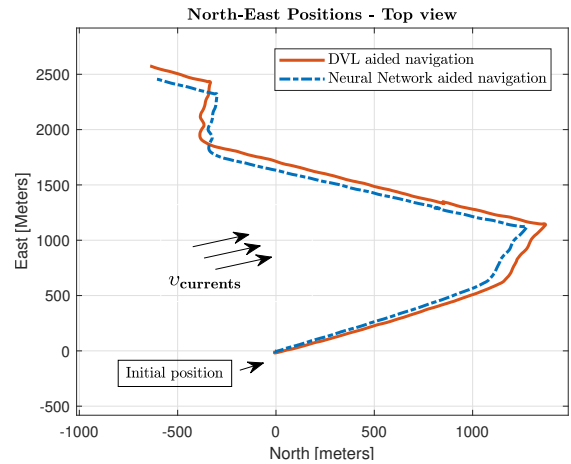}
    \caption{Neural network aided navigation - Top view}
    \label{fig:Tethys_DVL_vs_ANN}
\end{figure}

\begin{figure}[h]
    \centering
    \includegraphics[width=8.2cm,trim={0cm 0.3cm 0.3cm 0.3cm},clip]{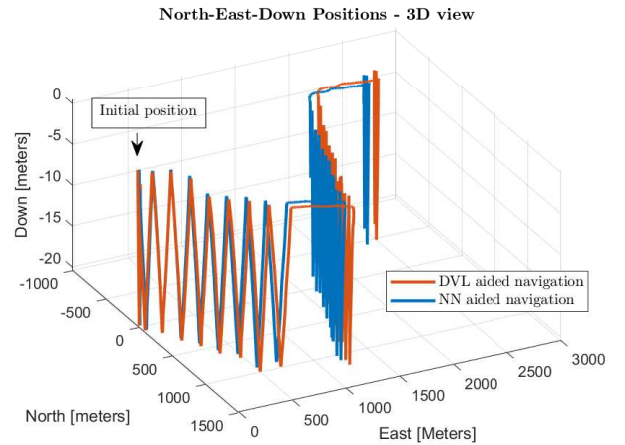}
    \caption{Neural network aided navigation - 3D view}
    \label{fig:Tethys_auv_3D}
\end{figure}

 Experimental data from the Tethys AUV was investigated in the initial study. Data from three individual surveys was concatenated as a time-series vector and used as training data. Datasets from another mission is used to test the neural network on unseen data. The duration of the test trajectory is approximately 3 hours long. The mission, illustrated in figure \ref{fig:Tethys_survey_monterey}, was conducted in shallow waters were the on-board Link Quest Micro DVL was able to get a \textit{bottom-lock}, although some samples were out of reach for the operating altitude of the DVL sensor. Outliners in the DVL data was removed and further filtered with Gaussian smoothing.   
\newline
The test trajectories consist of undulating saw-tooth motions with non-zero angle of attack $\alpha$ as showed in figure \ref{fig:Tethys_auv_3D}. Note that periodic GPS fixes was not accounted for in the results presented in figure \ref{fig:Tethys_auv_3D} and \ref{fig:Tethys_DVL_vs_ANN}.  
Two RNN networks was developed to isolate relative surge and sway predictions. The following table presents the training parameters used for the Tethys AUV.
 \begin{table}[h]
 \caption{ANN Training parameters - Tethys AUV}
    \centering
    \begin{tabular}{c|c}
    &    \\
    \hline
    Backpropogation Optimizer & Scaled Conjugate Gradient (SCG) \\
     MSE Surge Network & 0.0347 \\
       MSE Sway Network & 0.00588 \\
    Early Stopping Data Division & Randomly \\
    Early Stopping index & Training 70 $\%$, Val. 15$\%$, Test 15$\%$  \\
     Hidden layers & 3 \\
     Hidden neurons per layer & 40 \\
     Regressors per context layer & 5 \\
    \end{tabular}
    \label{tab:Tethys_ANN_parameters}
\end{table}
\newline
The results are presented in figure \ref{fig:Tethys_DVL_vs_ANN} and \ref{fig:Tethys_auv_3D}. The blue dotted line represents the predicted position based on estimated surge and sway velocities from the RNN network. The orange line is the estimated position based on measured DVL velocities.
\newline
A 3D view of the same trajectory is presented in figure \ref{fig:Tethys_auv_3D} where depth measurements from the pressure sensor is used for the vertical z-axis. 
\newpage
The positioning error between the predicted and DVL-aided horizontal positions was used to evaluate the performance. The positioning error is derived as
\begin{equation} \begin{bmatrix}
N_{error} \\
E_{error}
\end{bmatrix} = \begin{bmatrix}
\hspace{0.05cm} || \hspace{0.05cm} N_{est} - \hat{N} \hspace{0.05cm} || \hspace{0.05cm} \\
\hspace{0.05cm} || \hspace{0.05cm} E_{est} - \hat{E} \hspace{0.05cm}|| \hspace{0.05cm}
\end{bmatrix}
\end{equation}
Where $\hat{N}$ and $\hat{E}$ are ground truth north and east positions respectively. 
\begin{figure}[h]
    \centering
    \includegraphics[width=7.4cm,trim={0cm 0.1cm 0.3cm 0.3cm},clip]{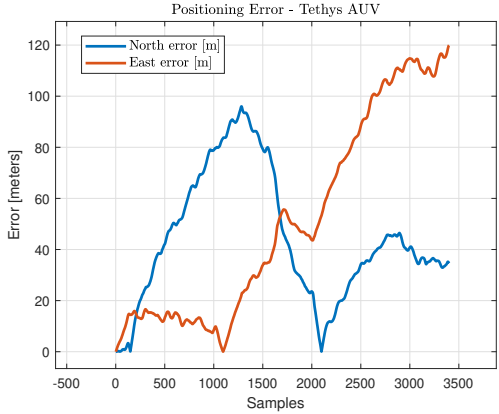}
    \caption{Positioning Errors - Tethys AUV}
    \label{fig:Tethys_auv_error}
\end{figure}
\newline
The positioning error relates to an approximate displacement of 2500 meters north and 1500 meter displacement in the east direction.
Note that the ground truth north and east positions estimated based on DVL velocities also have estimation errors with an 3-4 $\%$ navigational accuracy \cite{b2}. 
\subsection{Case Study 2 - Underwater Glider}
\begin{figure}[h]
    \centering
    \includegraphics[width=8.8cm,trim={0.1cm 0.2cm 0.5cm 0.2cm},clip]{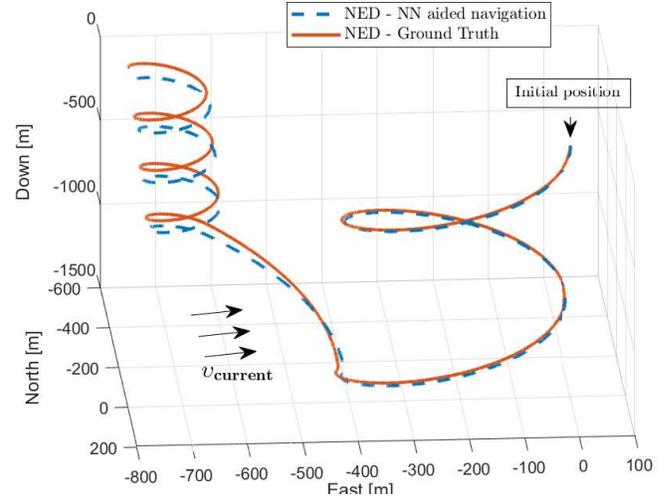}
    \caption{Estimated NED position vs ground truth}
    \label{fig:seawing_ned}
\end{figure}

\begin{figure}[h]
    \centering
    \includegraphics[width=8.5cm,trim={1cm 0.3cm 1cm 0.4cm},clip]{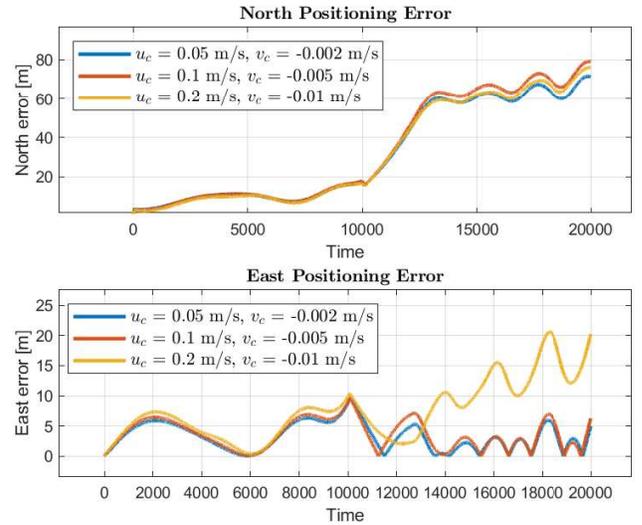}
    \caption{Positioning error with increasing ocean currents}
    \label{fig:seawing_ocean_error}
\end{figure}

A second case study was conducted with the simulated autonomous underwater glider. Training and test datasets were generated by the Simulink simulation of the glider dynamics. A two-dimensional ocean current model was added in the dynamics to create a simulated environment that reflects real ocean conditions. Simulated ocean currents are assumed to be constant, but time-varying during glider rotations.
\newline
Due to decoupled attitude and heading controllers from the simulated trajectories, the interaction between the surge and sway dynamics is assumed to be neglectable. Thus, two isolated RNN networks was developed to predict the relative surge and sway velocities.
\newline
Table \ref{tab:glider_ANN_parameters} presents the neural network training parameters for the two RNN networks.
  \begin{table}[h]
 \caption{ANN Training parameters - Glider}
    \centering
    \begin{tabular}{c|c}
    &    \\
    \hline
    Backpropogation Optimizer & Scaled Conjugate Gradient (SCG) \\
     MSE Surge Network & 0.000212 \\
       MSE Sway Network & 0.00000459 \\
    Early Stopping Data Division & Randomly \\
    Early Stopping index & Training 70 $\%$, Val. 15$\%$, Test 15$\%$  \\
     Hidden layers & 3 \\
     Hidden neurons per layer & 50 \\
     Regressors per context layer & 5 \\
    \end{tabular}
    \label{tab:glider_ANN_parameters}
\end{table}
\newline
The test trajectory presented in figure \ref{fig:seawing_ned} consists of two spiral motions and a wings-levelled movement. Three different ocean current scenarios was simulated with increasing magnitude, see figure \ref{fig:seawing_ocean_error}. The plot illustrated in figure \ref{fig:seawing_ned} shows the test dataset in presence of  low currents - $u_{c} = -0.05 \hspace{0.1cm} m/s$ and $v_{c} = -0.002 \hspace{0.1cm} m/s$. 
\newline
RNN velocity predictions is presented by the blue dotted line in figure \ref{fig:seawing_ned}. The orange line represents ground truth simulated NED positions. The glider trajectory was simulated for 2.7 hours. The remaining simulations with increasing ocean currents are presented in figure \ref{fig:seawing_ocean_error} where the north and east positioning errors are compared by the three different scenarios. Note that the x-axis relates to total samples with a rate of 2 Hz. Accordingly, the real simulation time was 10 000 sec.   
\newline
During the two first simulations with low and medium strong currents the positioning error is slightly increased. The simulation with strongest ocean currents induced a larger divergence for the east positioning error, while a low increase in the north error. 
    \section{Conclusions And Further Work}
    A neural network approach to aid dead-reckoning navigation for AUVs with a limited sensor suite was proposed in this work. Experimental data from an IMU, a pressure sensor and control actions were gathered from sea-trials and simulations with correlating ground truth DVL and simulated velocities. The objective for the trained RNN networks is to complement AUV navigation in absence of acoustic navigational instruments. Results from the proposed method show promising potential considering a limited sensor payload. Improvements can be made by re-initializing the DR algorithm with GPS fixes when the AUVs are surfacing. The positioning error for the underwater glider grows slightly with increasing magnitude of ocean current disturbances as illustrated in figure \ref{fig:seawing_ocean_error}. Glider positioning errors are significantly lower compared to DVL-less traditional navigation algorithms used in commercial gliders \cite{b41}. However, sensor noise and random walk errors were not present in the simulated IMU measurements. Further iterations of the simulated environment will focus on adding more realistic scenarios by introducing sensor errors, GPS fixes and vertical decomposition of the ocean currents.
    \newline
    \newline
    Recommendations for further work include investigating alternative methods to obtaining experimental data. Vision based pose estimation is considered a promising candidate which avoids replacing the DVL or acoustic modem with a "dummy" sensor. Another interesting subject is to extend the deep learning approach to other underwater robots like remotely operated vehicles (ROVs). Compared to under-actuated AUVs, small/miniaturized ROVs are easy to deploy and does not require large displacements to excite the ROV dynamics, making the experimental procedures less time-consuming.  
    \section{Acknowledgments}
    We thank the Monterey Bay Research Institute (MBARI) for granting access to mission data from the Tethys AUV.
    \newline
 This work was supported by the OASYS project funded by the Research Council of Norway (RCN), the German Federal Ministry of Economic Affairs and Energy (BMWi) and the European Commission under the framework of the ERA-NET Cofund MarTERA.

\end{document}